\documentclass[letterpaper, 10 pt, conference]{ieeeconf}

\IEEEoverridecommandlockouts                              

\usepackage{graphicx} 
\usepackage{amsmath} 
\usepackage{balance}
\usepackage{algorithm,algorithmic}

\title{\LARGE \bf
G-VOM: A GPU Accelerated Voxel Off-Road Mapping System
}

\author{Timothy Overbye and Srikanth Saripalli
\thanks{Timothy Overbye and Srikanth Saripalli are with the Department of Mechanical Engineering, Texas A\&M University, College Station, TX 77840, USA
        {\tt\small overbye2@tamu.edu}
        {\tt\small ssaripalli@tamu.edu}}%
}

\begin{document}

\maketitle
\thispagestyle{empty}
\pagestyle{empty}

\begin{abstract}

We present a local 3D voxel mapping framework for off-road path planning and navigation. Our method provides both hard and soft positive obstacle detection, negative obstacle detection, slope estimation, and roughness estimation. By using a 3D array lookup table data structure and by leveraging the GPU it can provide online performance. We then demonstrate the system working on three vehicles, a Clearpath Robotics Warthog, Moose, and a Polaris Ranger, and compare against a set of pre-recorded waypoints. This was done at 4.5~m/s in autonomous operation and 12~m/s in manual operation with a map update rate of 10~Hz. Finally, an open-source ROS implementation is provided.

https://github.com/unmannedlab/G-VOM

\end{abstract}

\section{INTRODUCTION}





Navigation is one of the fundamental problems in mobile robotics and cannot be accomplished without a good map. As vehicle speeds increase, both the update rate and fidelity of the map become more important. A good map lets us transform data from different sensors into a data structure that can be used for planning. Typically this can be broken down into two parts. First, obstacles are identified and the drivable area is separated from the non-drivable area. Second, factors such as terrain roughness or slope are identified and are used to define what parts of the free space are more or less preferred. Ultimately, this needs to be formatted into a structure that can be understood by the planner.

But what is a good map? There are two factors that could be said to determine the quality of a map. The fidelity, that is, how accurately does the map represent the real world. This is composed of resolution, obstacle detection robustness, drivable vs non-drivable surface segmentation, terrain roughness estimation, and other such things. Second is rate, that is, how fast can sensor data be taken in and turned into the map. The highest fidelity map in the world is of no use if the planner doesn't receive it in time.

   \begin{figure}[tpb]
      \centering
      \framebox{\parbox{3in}{
      
      \includegraphics[width=\linewidth]{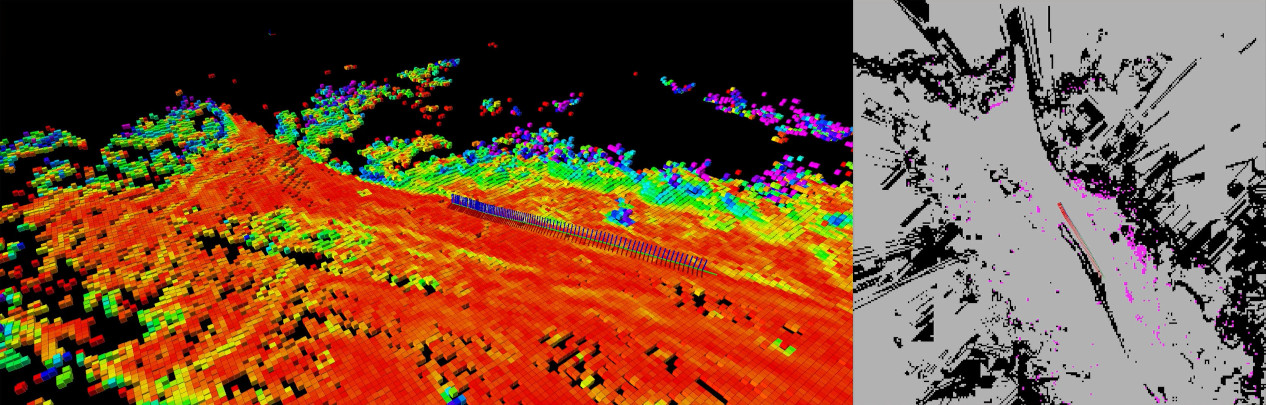}
      \includegraphics[width=\linewidth]{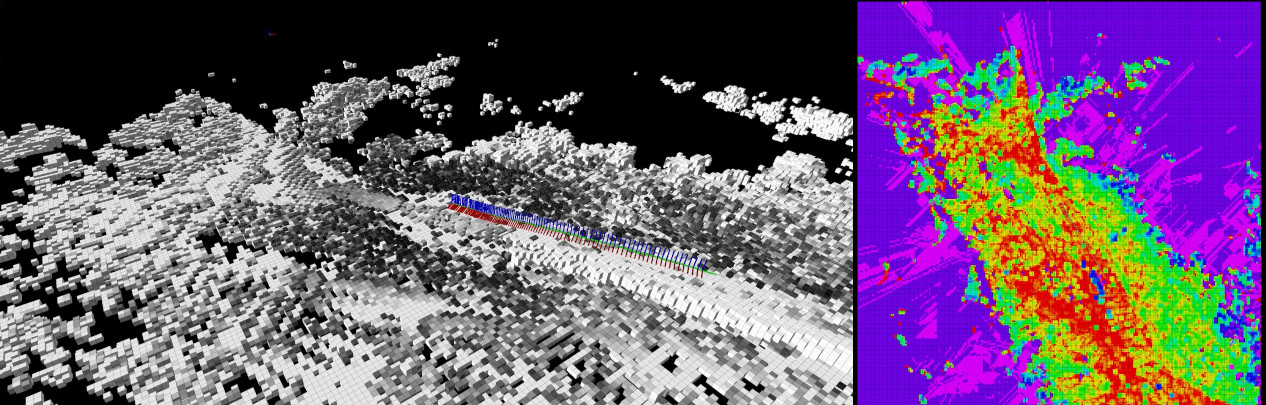}
	}}
      
      \caption{Outputs of the mapping process. From top left clockwise, surface height (red flatter slope, green steeper slope), obstacle map (hard obstacle in black and soft obstacles in pink), terrain roughness map (red less rough, green more) and negative obstacles (pink), and voxel map (lighter more opaque, darker more transparent). }
      \label{fig:outputs}
   \end{figure}
   
Fidelity is of particular importance for off-road navigation due to the environment's unique considerations. While roads can be safely assumed to be drivable, off-road surfaces cannot always be. For example, there is a high risk of getting stuck if driving though mud. However, mud will often look like flat, drivable ground to sensors. Off-road operation also has a more complex obstacle set. Both a tree and a bush are obstacles yet, if necessary, the bush could be driven though while the tree could not. We refer to these as hard and soft obstacles. Even within soft obstacles we might want to know their density to minimize the vegetation driven through. Finally, the interactions between the vehicle and the ground must be considered. Rough ground could cause unacceptable vibrations, a steep slope could roll the vehicle, and wet grass might change the stopping distance. We don't solve all these cases but they must be considered while working on off-road navigation.

However, both fidelity and rate require computation time. An increase in one often leads to a decrease in the other. Even with increasingly fast hardware and new optimization techniques, trade-offs must be made. Voxels provide a good middle ground between these two objectives. Due to their 3D nature, they contain more information about the world than the more common 2D or 2.5D maps while still being fast. Additionally, due to their grid nature, most operations on a voxel map are easily parallelizable. This makes them a prime candidate for calculation on a GPU yielding further rate increases.

In this work we present G-VOM, a GPU accelerated voxel mapping system for off-road navigation. Our system requires only odometry and lidar sensor data. This data is stored in a space efficient voxel map where a drivable surface height map, positive and negative obstacle maps, terrain slope maps, and terrain roughness maps are computed (figure~\ref{fig:outputs}). These can then be exported as a series of 2D maps for easy integration with existing planners. An implementation of G-VOM is provided as an open-source ROS package\footnote[1]{https://github.com/unmannedlab/G-VOM}.

\section{RELATED WORK}

The most common mapping approach for mobile ground robots has been the occupancy grid~\cite{probobalistic_robotics}\cite{occupancy_grid}\cite{grid_map_1}. This is a 2D map, typically centered on the robot, where each pixel in the map contains the probability of that pixel being occupied. These probabilities come from sensor measurements (initially planar lidar but other sensors, such as stereo cameras or 3D lidar, are now used) either taken during runtime or loaded from saved data. 

However, as environments become more complicated the limited data in a 2D occupancy map becomes insufficient to make safe choices. Negative obstacles such as holes or ditches don't fit nicely into categories of occupied or empty. Additionally, features of the terrain itself such as slope or roughness can have large effects on traversability but shouldn't always be considered as obstacles. Finally, in the off-road environment, there are soft obstacles such as bushes or tall grass which may be avoided most of the time but could be driven through if needed. This is in contrast to the hard obstacles such as trees, rocks, or buildings that can never be driven through.

A natural extension of the occupancy grid is to have multiple different grids, each corresponding to a different classification of obstacle or sensor type. This is commonly referred to as a layered costmap and is widely used~\cite{multilayer_obs}\cite{multilayer_sensors}\cite{my_planning_paper}. When one of these layers is the terrain height it's typically referred to as a 2.5D map. However, once non-obstacle layers, such as slope or roughness, are added the selection of a suitable cost function becomes an important consideration.

As this cost function can be difficult to properly define, there has been recent work on once again simplifying maps to only two states, traversable or non-traversable. This is typically accomplished via machine learning methods~\cite{drivable_learning_1}\cite{drivable_learning_2}. A classifier will be given either raw sensor data or data from the previously mentioned map layers. It can then be trained either with pre-labeled datasets or can learn traversable areas by driving though them. 

Although these methods can all provide very good solutions, they are still fundamentally 2D and are thus limited in the data that can be stored in each pixel. Overhangs or tree limbs being one example, they can either be classified as an obstacle or forgotten but there is not a natural way to represent them~\cite{multilevel_surface}. A better way to represent them is to extend the map into 3D. If we extend the occupancy grid into 3D we get the voxel map~\cite{voxel_1}\cite{voxel_occupancy}. 

Unfortunately, while this additional dimension is helpful it also greatly increases the amount of information that must be stored. This can quickly become unreasonable as more data is added to each voxel or if multiple grids are to be kept in a buffer.

A popular solution to this problem is the octree~\cite{octree}. In an octree 3D space is broken up into octants. If an octant does not contain empty space it is then broken up into 8 child octants. If an octant does only contain empty space it is marked as empty and not expanded. This process continues until the desired resolution is reached. Since most space within the map will be empty, octrees can yield significant memory savings as voxel data is only stored in the final leaf nodes. For this reason this structure is popular for 3D voxel systems such as Octomap~\cite{octomap}. There are also may different extensions of octrees that can provide application specific benefits~\cite{voxel_occupancy}\cite{jittree}\cite{octree_gpu}.

However, there are still trade-offs with octrees. First, since empty space is represented as a childless node, there is no way to store information about empty space. Secondly, the space efficiency of the octree comes at the expense of access time efficiency. In a 3D grid the time complexity of random access is $O(1)$. However, an octree has a random access time complexity of $O(log(d))$ where $d$ is the depth of the tree.

Another method consists of creating a 2D grid where each element in the grid is a list of voxels~\cite{multilevel_surface}. Each of these lists starts at the lowest occupied voxel in the column and ends at the highest. Since most data of interest will be near the ground this method can also produce significant space savings. However, it also has a time complexity trade-off with a complexity of $O(n)$ with $n$ being the length of the list.

Beyond memory concerns, the added dimensionality of 3D maps can also make planning more difficult. Therefore, most 3D mapping systems for ground vehicles will use the 3D map as a source to make 2D maps which are then sent to the planner~\cite{skimap}\cite{virtual_surfaces}. When there may be multiple drivable levels within a single map (like in a stairwell) multi-level maps are often used. These produce a 2D map for each extracted drivable region along with information about where levels connect.

Our approach uses a 3D voxel map with a fixed map size centered on the vehicle. The map is stored as a 3D array lookup table with a smaller 1D data array. Finally, the voxel map is used to make a series of 2D maps which can be sent to the planner.

\section{APPROACH}

\subsection{Map Storage}
The first thing that must be considered in implementation of a voxel map is the data structure used to store the map. Since we use a fixed map size, typically $[256,256,64]$ with a resolution of 40~cm, a 3D array would be a simple solution. However, this would take up too much memory. Instead the data of the map consists of two arrays. First is the lookup table, a 3D array the size of the map. Second, the data array, a 1D array who's length is the number of occupied voxels in the map and contains metrics of the occupied voxels. The metrics for each voxel are number of returns within the voxel, number of rays passing though the voxel, and the height of the lowest return within the voxel. If a voxel at some index $[x,y,z]$ is occupied then the lookup table array contains that voxel's index in the data array. If the voxel is not occupied then the value of the lookup table at that index is $-1 - N_m$ where $N_m$ is the number of "misses", that is, lidar rays that passed though the voxel but did not end in it in. Stored along with each map is the coordinates of that map's origin in world space. This origin is always an integer multiple of the map resolution such that voxels will always line up across maps. 
This structure allows us to keep some information about empty voxels while still saving space compared to a single 3D array. Additionally, it maintains $O(1)$ random access time at the cost of a more complex map creation process.

\subsection{Processing Outline}

\begin{figure}[htbp]
      \centering
      \framebox{\parbox{3in}{
      
      \includegraphics[width=\linewidth]{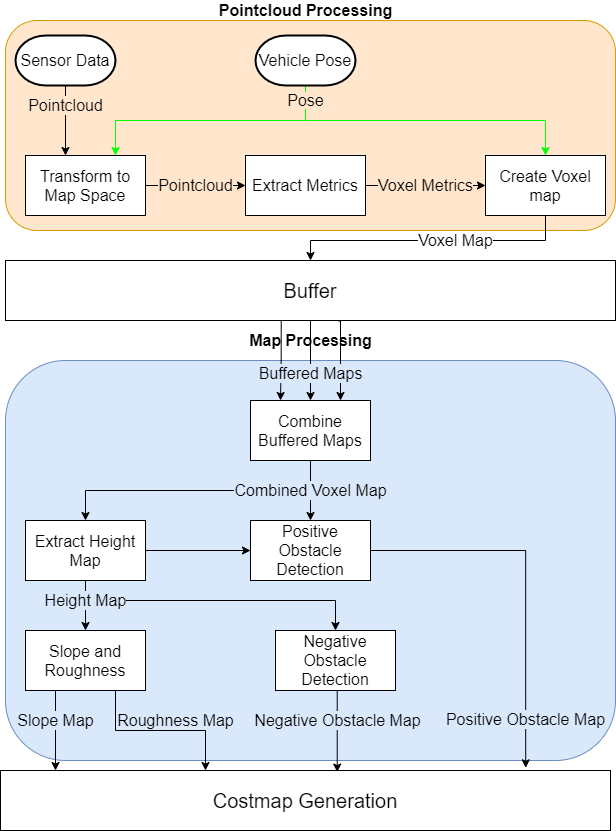}
	}}
      
      \caption{Overview of the local mapping process. Top shaded region is pointcloud input and processing. Bottom shaded region is combined map processing. }
      \label{fig:process_overview}
\end{figure}

The mapping process itself can be broken into two parts, pointcloud processing, and map processing. Figure~\ref{fig:process_overview} provides an overview of the process. Pointcloud processing takes the pointcloud and odometry data and creates a voxel map which is then added to the buffer. Map processing combines all the maps in the buffer into a single map and produces the output maps. Both of these processes can be run asynchronously from each other. Additionally, there is no limit to the number of different pointcloud processing nodes. This enables multiple different lidars to feed into the same map.

\subsection{Pointcloud Processing}
Once the pointcloud is received it is transferred to the GPU along with the relevant odometry data. From here all pointcloud operations are computed with one thread per point in the cloud. The odometry data then is used to transform the pointcloud into the map frame. Two passes are then done over the pointcloud. The first pass counts the number of unique occupied voxels and creates the lookup table array and data array. In the second pass the lidar rays are traced and metrics such as the number of hits and misses per voxel and minimum return height are calculated. This data is put into the data array and the completed map (lookup array, data array, map origin) is inserted into the buffer.

\subsection{Map Processing}

Map processing begins by combining all the maps in the buffer into a single voxel map. This is done by offsetting each of the map indices by the offset between the buffer map and the combined map. The combined map uses the location of the most recent buffer map as it's origin. Voxel metrics are then combined with number of hits and misses being added together and minimum return heights compared and the minimum taken. These operations are done with one thread per voxel in the combined map. The following 2D maps are computed with one thread per pixel.

From this voxel map we extract a height map over the lowest surface on the map. That is, for each pixel in the height map, its value is calculated by taking the height of the minimum height return of the minimum height voxel within each column. We assume that this height map represents the height of the driveable surface and that the vehicle is constrained to this surface. 

We then use this assumption and map to calculate positive obstacles. For each pixel in the height map we check above it for any voxels of a height between then minimum obstacle height and maximum obstacle height. If there are voxels in that range then that pixel is marked as a positive obstacle in the positive obstacle map. The density of the obstacle is the weighted average density of all voxels in the obstacle range. This map is further thresholded to produce the hard and soft obstacle maps. With hard obstacles having a density above the threshold and soft obstacles below.

Next are calculated the slope and roughness maps. The slope of each pixel is calculated by least squares fitting of a plane taking an NxN square of pixels around the pixel of interest. The roughness of that pixel is the average squared error. We are only concerned with roughness of a frequency lower than that of a pixel so we don't consider the roughness within a pixel. Since our pixels (40~cm by 40~cm) are comparable in size to the vehicle's tires we assume any higher frequency roughness will be filtered by the tires.

\begin{figure}[htbp]
      \centering
      \framebox{\parbox{3in}{
      
      \includegraphics[width=\linewidth]{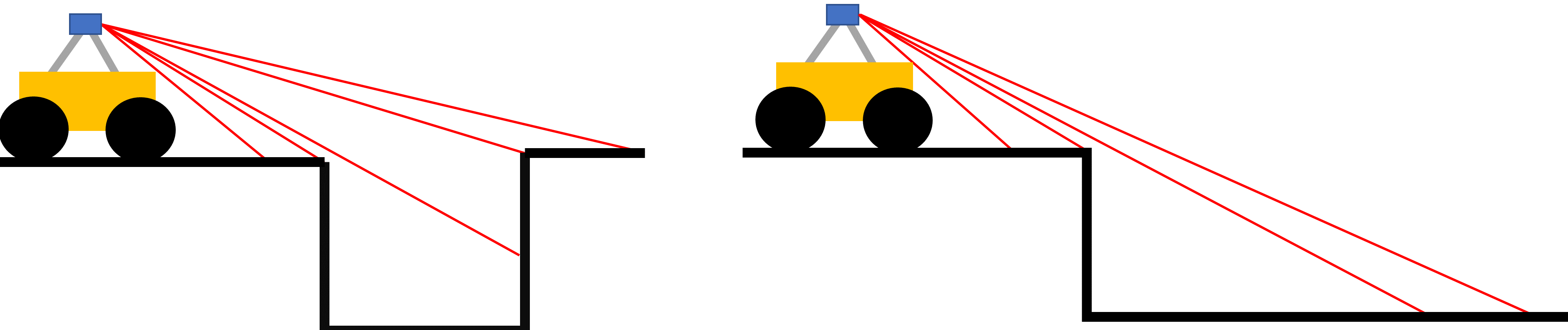}
      \includegraphics[width=\linewidth]{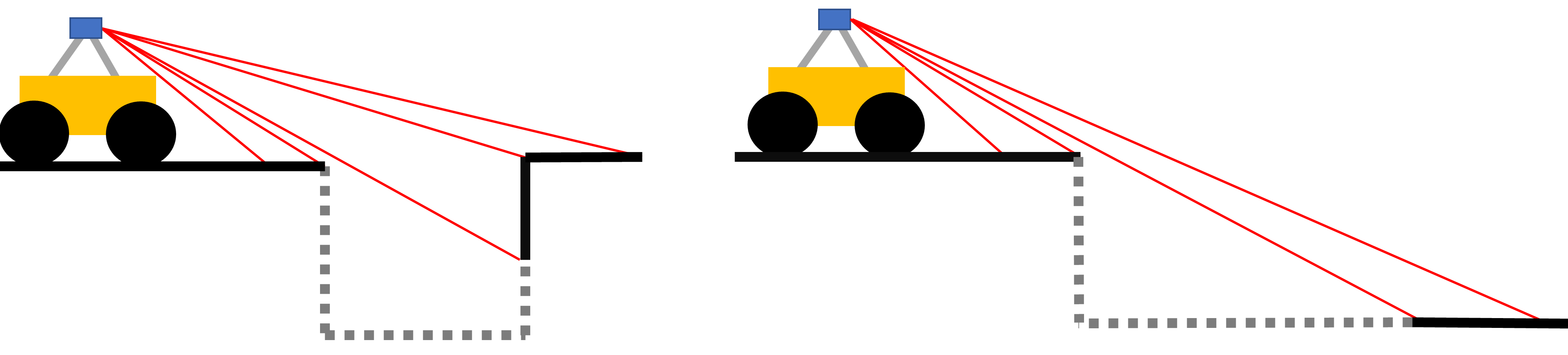}
      \includegraphics[width=\linewidth]{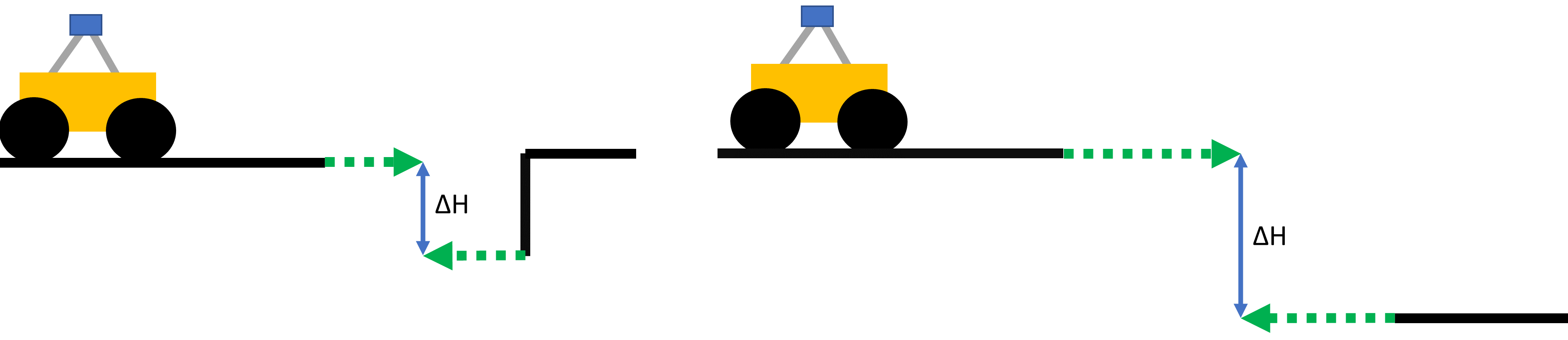}
	}}
      
      \caption{Types of negative obstacles. Left hole type, right cliff type. Top ground truth, middle viable (solid) and obscured (dashed) surfaces, bottom assumed surfaces (dashed arrows) and assumed height difference ($\Delta$H). }
      \label{fig:neg_obs_process}
   \end{figure}

The final one is the negative obstacle map. We assume two basic types of negative obstacles, holes and cliffs (figure~\ref{fig:neg_obs_process}). The fundamental difficulty in detecting negative obstacles is that they can't be directly seen. Rather, they must be inferred from surrounding data. If the ground can be seen we don't treat it as a negative obstacle, rather slope information is used to determine traversability. Our system is based on the assumption that unobserved terrain continues at the same height as the last observed terrain in that direction. Figure~\ref{fig:neg_obs_process} shows this assumption with the green dashed arrows representing the assumed terrain. If the height difference between the assumed heights ($\Delta H$ in figure~\ref{fig:neg_obs_process}) is greater than the negative obstacle threshold the point is marked as a negative obstacle.

\begin{figure}[htbp]
      \centering
      \framebox{\parbox{2.4in}{
      
      \includegraphics[width=\linewidth]{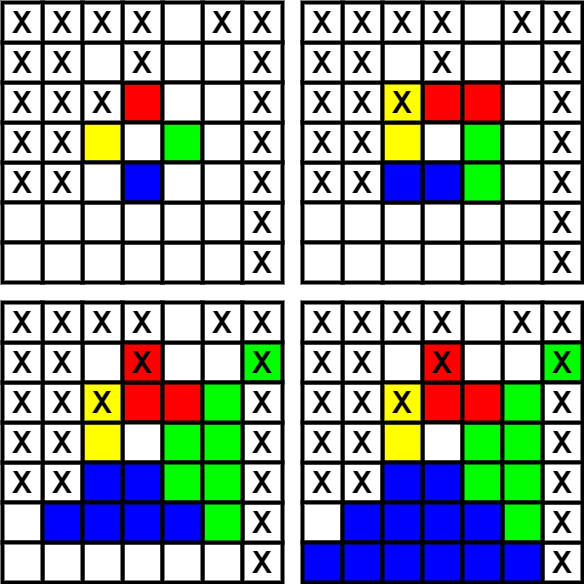}
	}}
      
      \caption{Search process used to find defined surfaces for negative obstacle detection. Center pixel is start pixel, "X" represents known ground, colors represent search areas. Top left, initial state. Top right, search expanded until yellow finds known ground. Bottom left, expanded until red and green find known ground. Bottom right, expanded until blue reaches edge of grid. In this case, the cells found by yellow, red, and green would be used.  }
      \label{fig:neg_obs_search}
   \end{figure}

For the actual implementation, we assume that negative obstacles only exist in areas of the map where the height map hasn't been defined. For each pixel in the undefined region we search in each direction in a cone shape until a defined surface has been found or until a maximum search distance is reached. This process is shown in figure~\ref{fig:neg_obs_search}. The maximum height between each of the defined surfaces is then calculated. If the maximum difference between any of these assumed height is larger than the negative obstacle threshold that pixel is defined as a negative obstacle. In the example the height values from the yellow, red, and green cells would be used.

Finally, each of these maps are published and can then be turned into a costmap by the planning system.

\section{RESULTS}

The system was implemented on three vehicles that were tested at the Texas A\&M RELLIS campus. A Clearpath Robotics Warthog\footnote[2]{https://clearpathrobotics.com/warthog-unmanned-ground-vehicle-robot/} and Moose\footnote[3]{https://clearpathrobotics.com/moose-ugv/}, and a Polaris Ranger\footnote[4]{https://ranger.polaris.com/en-us/} (figure~\ref{fig:vehicles}). 

\subsection{System Overview}

\begin{figure}[htbp]
      \centering
      \framebox{\parbox{3in}{
      
      \includegraphics[width=\linewidth]{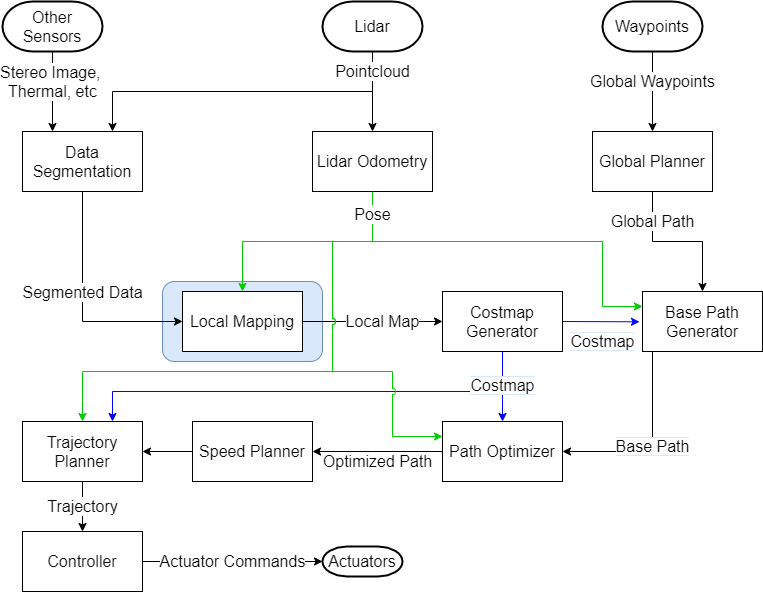}
	}}
      
      \caption{Overview of the test system, highlighted region is this paper's contribution. }
      \label{fig:system_overview}
\end{figure}

   \begin{figure*}[htbp]
      \centering
      \framebox{\parbox{7in}{
      
      \includegraphics[width=\linewidth]{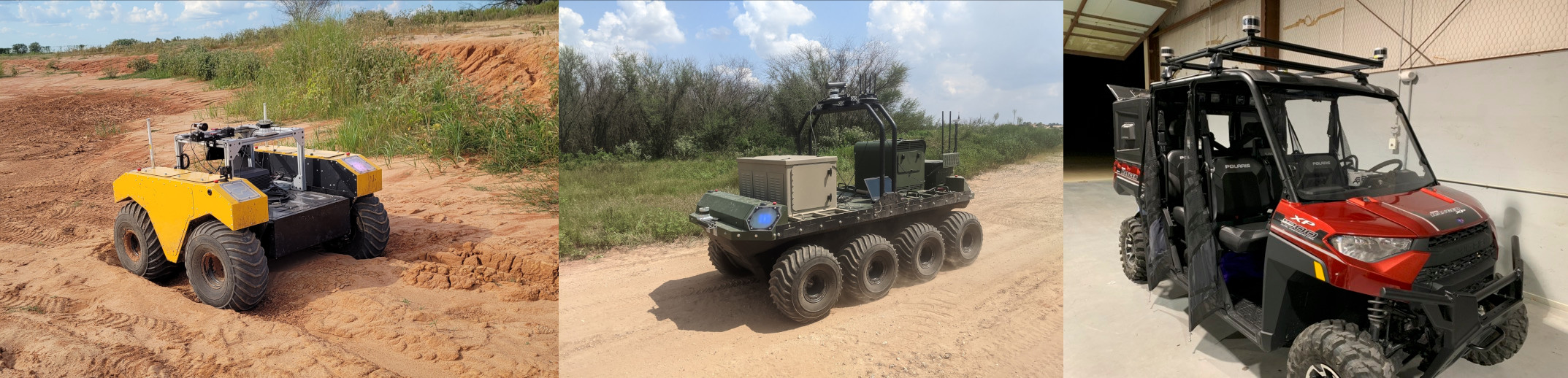}
	}}
      
      \caption{Our test vehicles, from left to right Clearpath Robotics Warthog, Moose, and Polaris Ranger.}
      \label{fig:vehicles}
   \end{figure*}
   
Our primary test vehicle was the Warthog (figure~\ref{fig:vehicles} left).  Sensing was done with a Ouster OS1-64 lidar, odometry was calculated using A-LOAM, an implementation of~\cite{loam}. Planning was done with the systems developed in~\cite{my_planning_paper}\cite{my_optimization_paper}. Basically, to create the costmap, each of the output maps get some weight assigned to them and the resulting per pixel sum is the cost in that pixel. A* is used to make a plan from the vehicle's location to the nearest waypoint. The A* path is then optimized~\cite{my_optimization_paper} and used as an input to a trajectory generator. The trajectories are made by simulating over a set of possible steering inputs with the lowest cost trajectory chosen. Finally, to follow the trajectory, we used an ILQR controller \cite{ilqr}. Figure~\ref{fig:system_overview} shows an overview of the software architecture with this work's contribution highlighted.

   \begin{figure}[htbp]
      \centering
      \framebox{\parbox{3in}{
      
      \includegraphics[width=\linewidth]{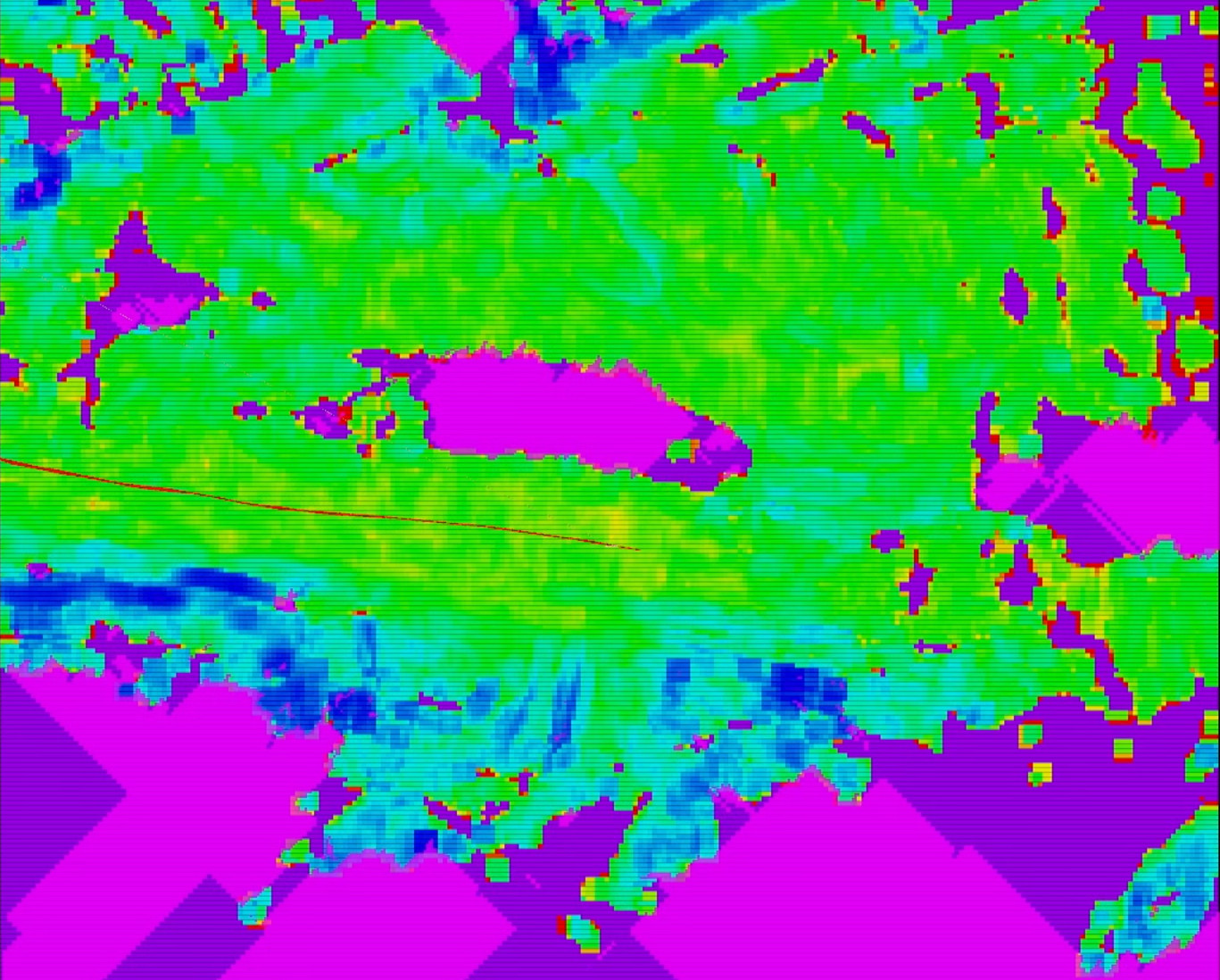}
	}}
      
      \caption{Roughness (yellow-green low, blue high) and negative obstacle (bright pink) maps produces on the ranger. Red line is the driven path, the path was driven above the step in figure~\ref{fig:hill_climb} from right to left. The negative obstacle in the middle of the figure corresponds to the lower region in figure~\ref{fig:hill_climb}.}
      \label{fig:ranger_map}
   \end{figure}

Testing was also done on the Moose and Ranger (figure~\ref{fig:vehicles} center and right). Both used an Ouster OS1-128 for sensing. No planning or control was done either. Figure~\ref{fig:outputs} shows the maps outputted while running the system on the Moose and figure~\ref{fig:ranger_map} shows the roughness and negative obstacles maps while running on the Ranger. The moose was driven at a max speed of 8~m/s and the Ranger at 12~m/s.

For all vehicles, the system was ran on a laptop with an Nvidia Quadro RTX 4000 GPU and an Intel i9-10885H CPU. On all vehicles our system ran with a mapping rate from 9-12~Hz.

\subsection{Experimental Results}

Here we present four test cases with the Warthog. Two long paths following a set of GPS waypoints, a short  uphill climb, and a short downhill climb.
The GPS waypoints for tests 1 and 2 were gathered by manually driving the vehicle over the path. The resulting waypoints were then downsampled such that the resulting waypoints were about 50-80~m apart. The starting and ending waypoints were kept fixed. The first two tests were driven at 4.5~m/s and the last two at 3~m/s.

   \begin{figure}[htbp]
      \centering
      \framebox{\parbox{3in}{
      
      \includegraphics[width=\linewidth]{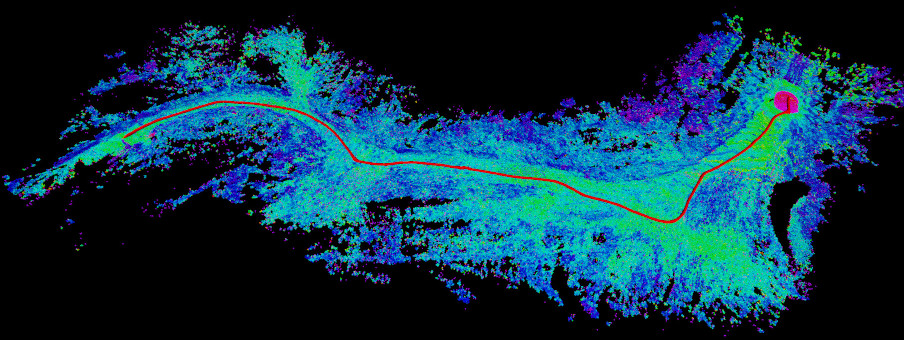}
	}}
      
      \caption{Test 1 showing the driven path over the full map. Color is roughness (lighter is less rough, darker more), red line is driven path.}
      \label{fig:path1}
   \end{figure}
   
\begin{figure}[htbp]
      \centering
      \framebox{\parbox{3.35in}{
      \includegraphics[width=\linewidth]{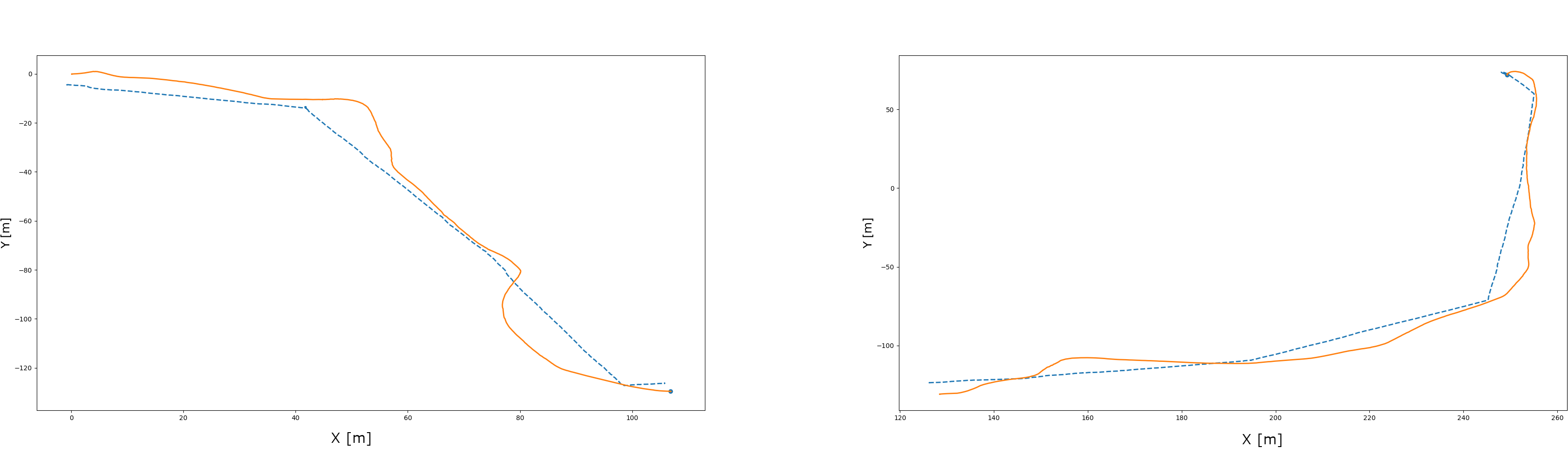}
      \includegraphics[width=\linewidth]{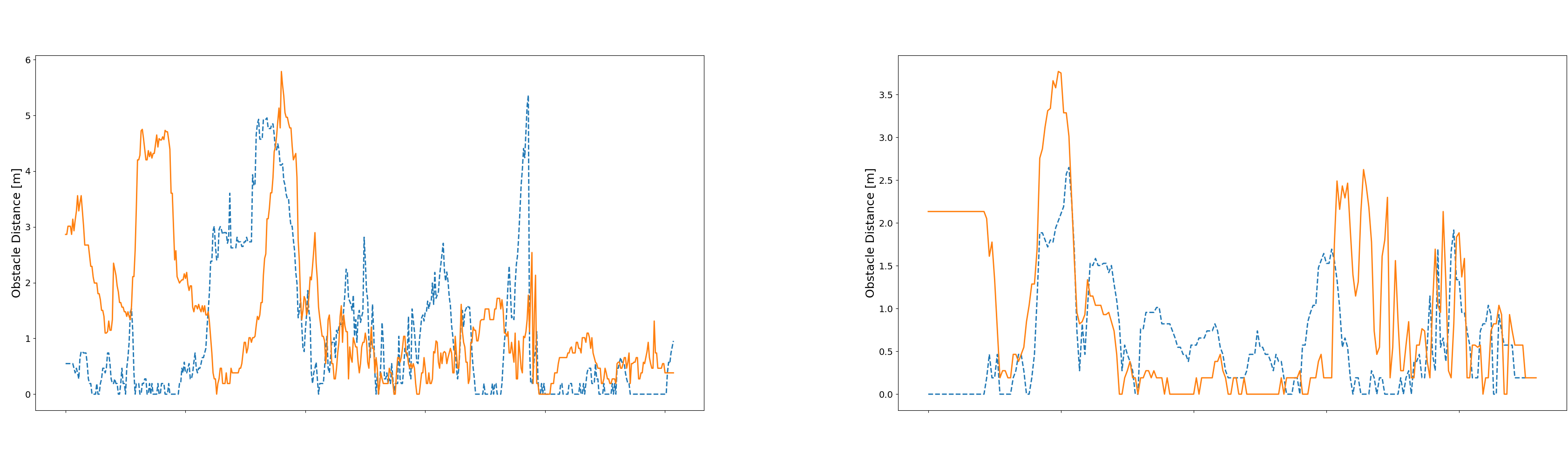}
      \includegraphics[width=\linewidth]{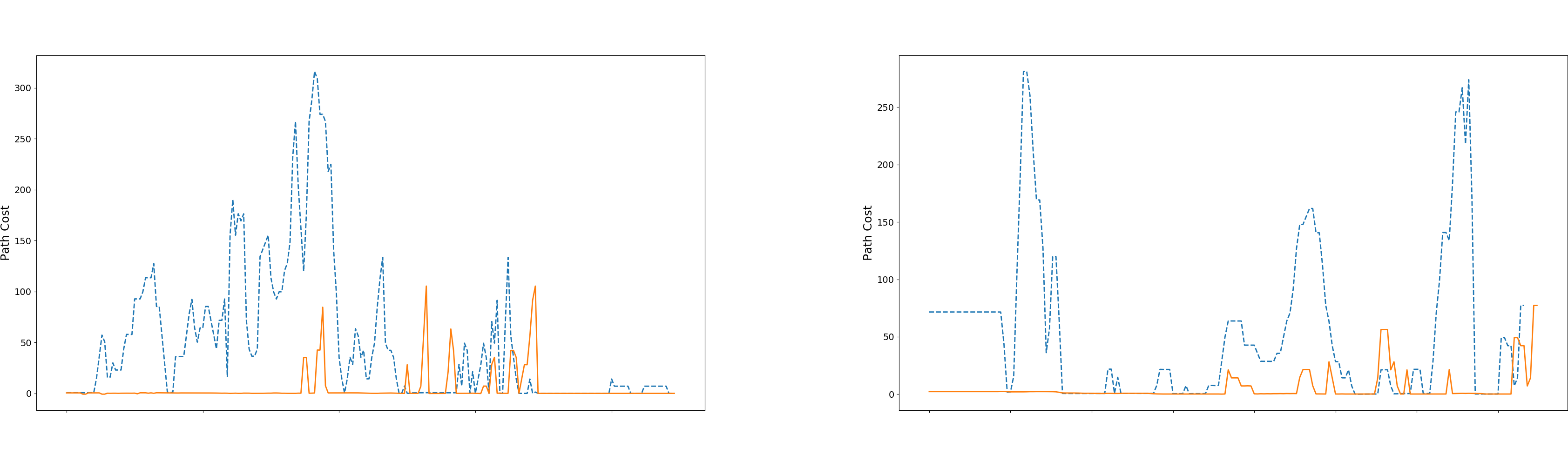}
	}}
      
      \caption{Test 1 (left) and 2 (right) results. Top is the driven path (solid orange) and the path formed by the waypoints (dashed blue). Middle is minimum distance to obstacle vs path length for both driven (solid orange) and waypoint (dashed blue). Bottom is the path cost vs length for both driven (solid orange) and waypoint (dashed blue). }
      \label{fig:path_results}
   \end{figure}

Figure~\ref{fig:path1} shows the driven path in red over the recorded map. Note that the shown map is a composite of the maps recorded during the test.  Figure~\ref{fig:path_results} (top left) shows the driven path overlaid on the path formed by the waypoints. Figure~\ref{fig:path_results} (middle and bottom left) shows a comparison of both the minimum distance between the path and obstacles and the instantaneous path cost as defined by the planner's cost function. Note that overall the driven path maintains a better distance from obstacles. Although both paths occasionally pass too close to obstacles. However, the driven path has a significantly lower cost compared to the waypoint path. This should come as no surprise as this cost is precisely what the planner is trying to minimize.

\begin{figure}[htbp]
      \centering
      \framebox{\parbox{3in}{
      
      \includegraphics[width=\linewidth]{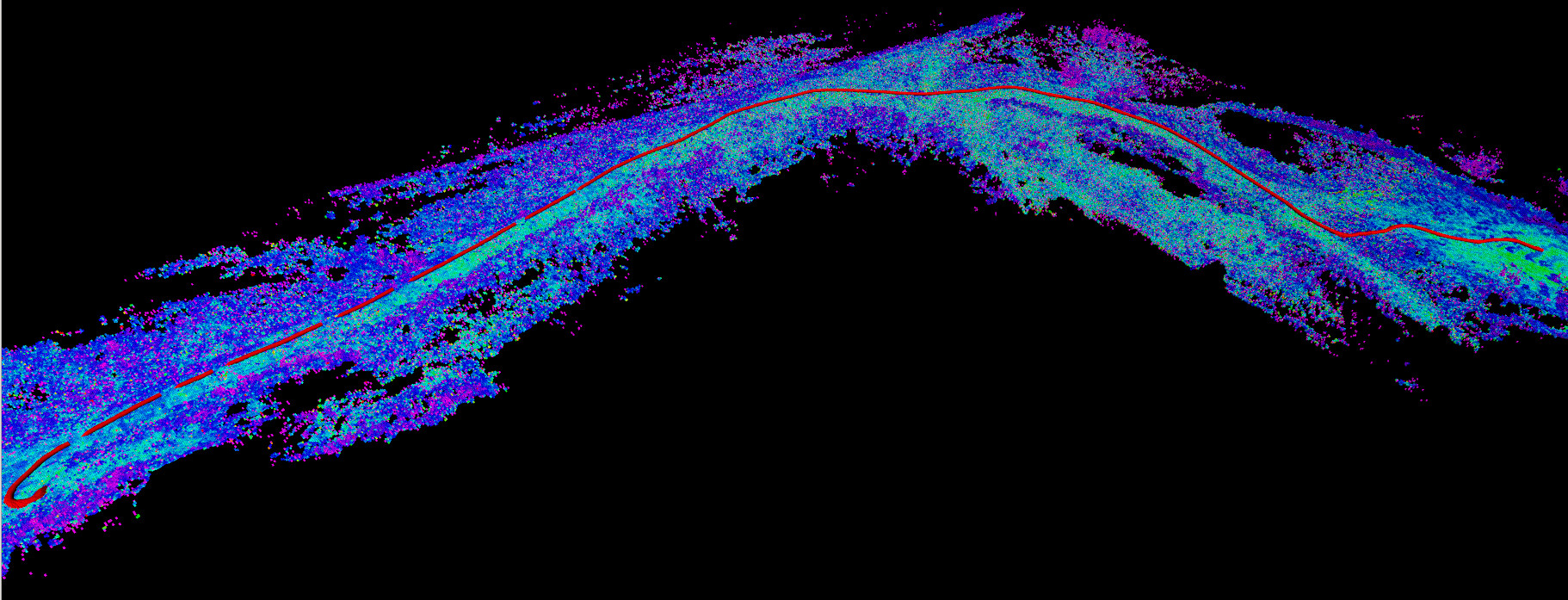}
	}}
      
      \caption{Test 2 showing the driven path over the full map. Color is roughness (lighter is less rough, darker more), red line is driven path.}
      \label{fig:path2}
   \end{figure}

The second test is similar to the first. Once again, figure~\ref{fig:path2} shows the driven path over the recorded map. In figure~\ref{fig:path_results} (top right) we can see that the driven path was closer to the waypoints than in the first test. Although there's still deviations. Much like the first test the driven path overall has a better distance from obstacles (figure~\ref{fig:path_results} right middle). However, during the middle section the driven path is closer to obstacles than the waypoint path. When looking at the instantaneous path cost (figure~\ref{fig:path_results} right bottom) we can see that some of these close obstacle encounters don't have a corresponding spike in the path cost. This is likely due to the costmap generator filtering out "single pixel" obstacles assuming they're just noise in the map.

	\begin{figure}[htbp]
      \centering
      \framebox{\parbox{3in}{
      
      \includegraphics[width=\linewidth]{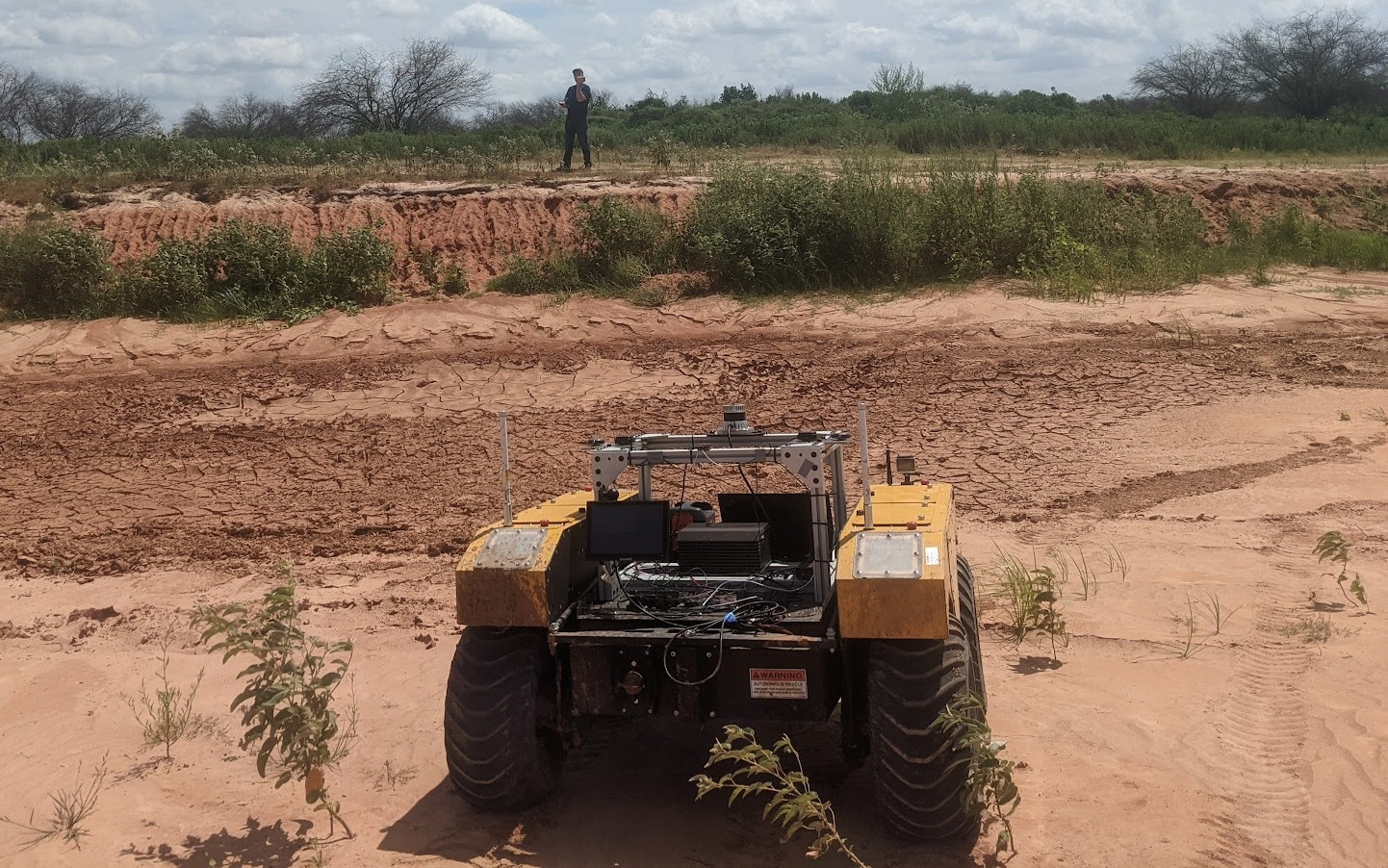}
	}}
      
      \caption{Step climb test area for Test 3 and 4. The vehicle is at the downhill start and end position. Person is at the uphill start and end position.}
      \label{fig:hill_climb}
   \end{figure}
   
	\begin{figure}[htbp]
      \centering
      \framebox{\parbox{3in}{
      
      \includegraphics[width=\linewidth]{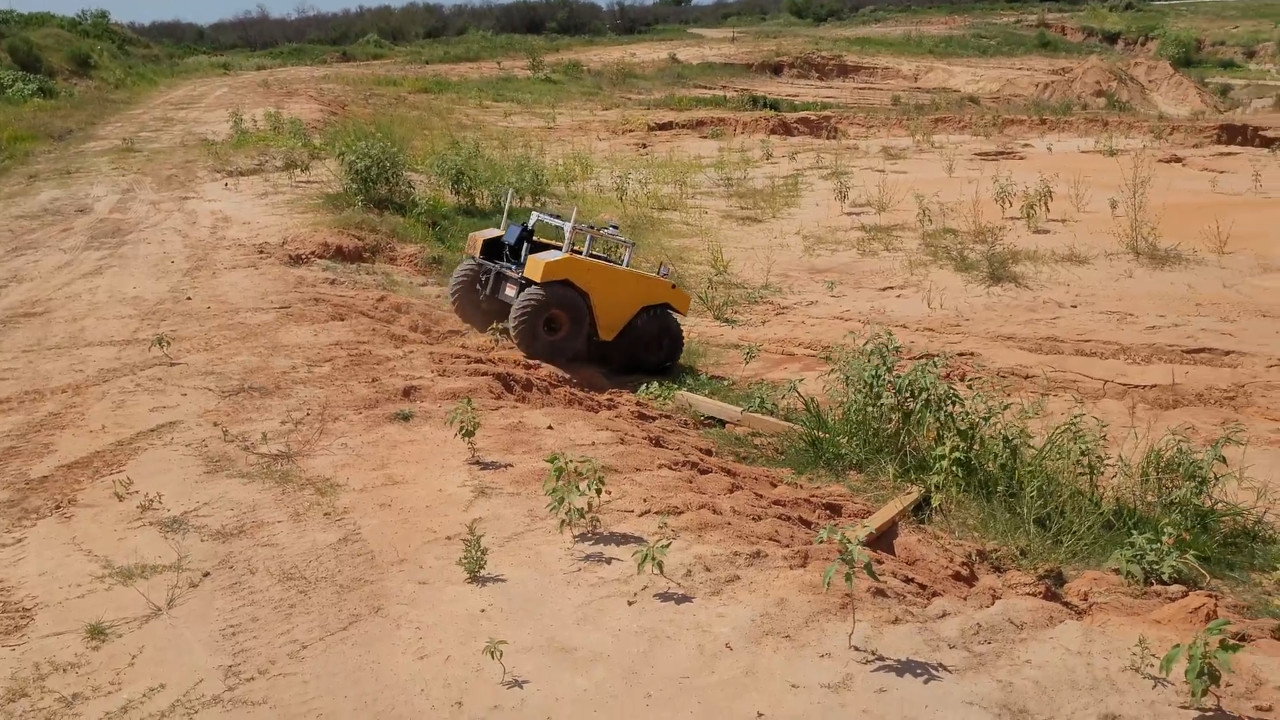}
	}}
      
      \caption{Step climb test area for Test 3 and 4. View from the top, the vehicle is climbing down the step.}
      \label{fig:hill_climb_2}
   \end{figure}
   
\begin{figure}[htbp]
      \centering
      \framebox{\parbox{3in}{
      \includegraphics[width=\linewidth]{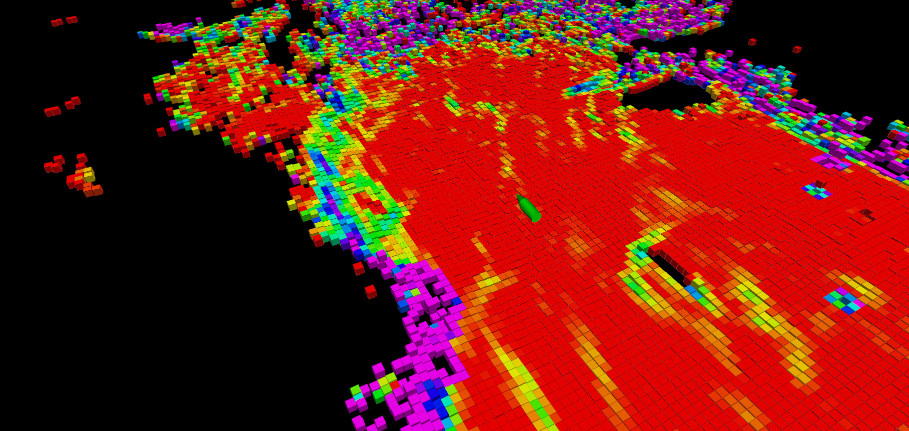}
      \includegraphics[width=\linewidth]{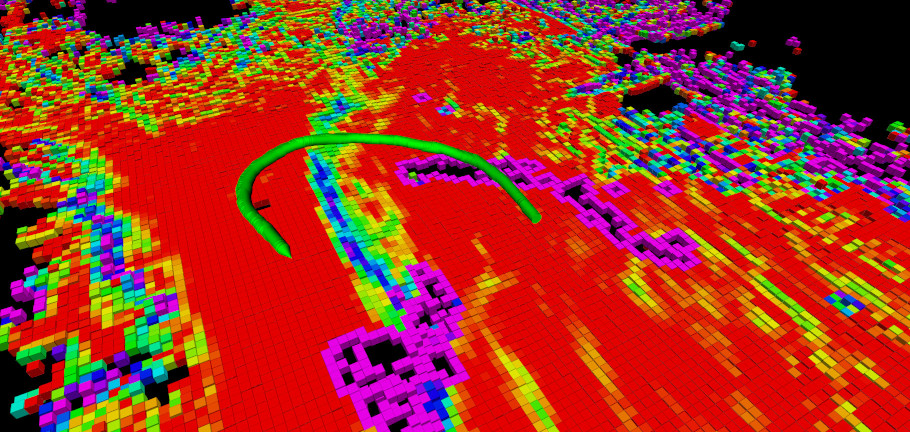}
	}}
      
      \caption{Test 3, uphill climb. Top is the initial position and map, bottom is final path (green) and map. Map color is slope (red is flat, blue is steep).}
      \label{fig:uphill}
   \end{figure}

The next two tests show the vehicle climbing up and down a small step, about 1.5~m, in the terrain. Figures~\ref{fig:hill_climb}~and~\ref{fig:hill_climb_2} show the test area. In the figure the vehicle is at the "downhill" waypoint and the person is standing at the "uphill" waypoint. The first of these tests has the vehicle climbing up the step. Figure~\ref{fig:uphill} shows the initial map on top and the driven path on the bottom. The point at which the vehicle climbs the hill is near the right border in figure~\ref{fig:hill_climb}. Note that the mapping system successfully creates a good slope estimation even with the vegetation. Additionally, the vehicle was successfully able to navigate to a lower slope portion of the step before turning around to reach the goal point.   
   
\begin{figure}[htbp]
      \centering
      \framebox{\parbox{3in}{
      \includegraphics[width=\linewidth]{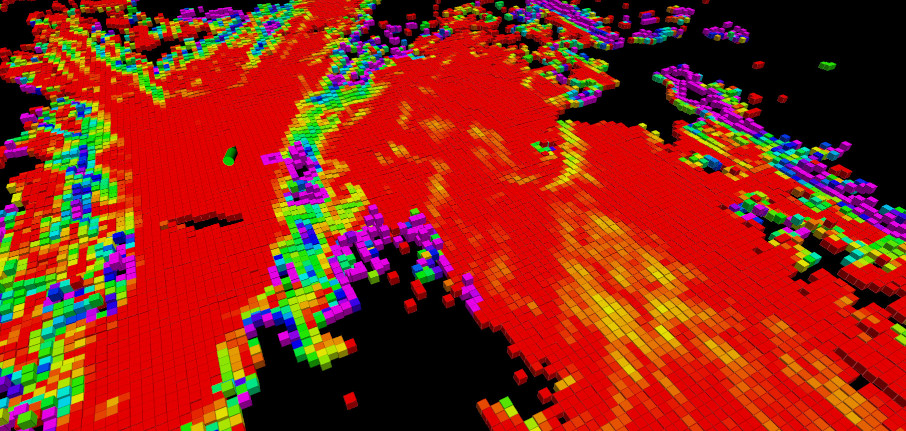}
      \includegraphics[width=\linewidth]{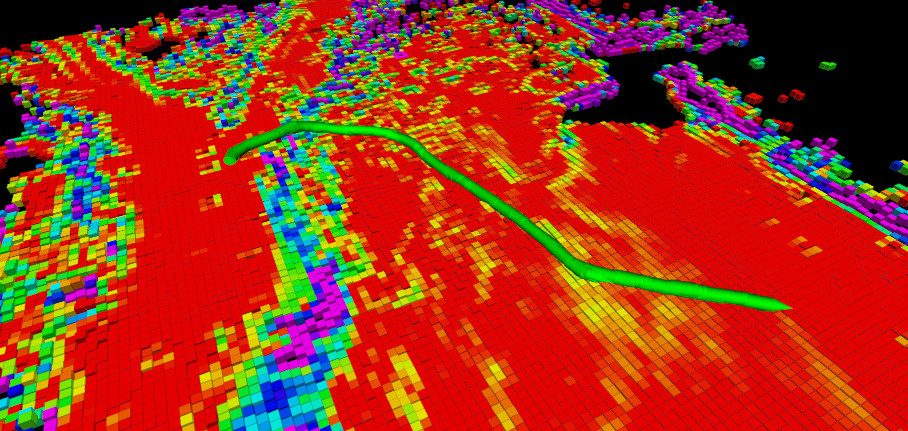}
	}}
      
      \caption{Test 4, downhill climb. Top is the initial position and map, bottom is final path (green) and map. Map color is slope (red is flat, blue is steep).}
      \label{fig:downhill}
   \end{figure}

The downhill climb is similar, initially there is a large obscured area near the ridge (figure~\ref{fig:downhill} bottom center in top image). This area is marked as a negative obstacle until further information about the lower ground is provided. However, the driven path avoids this area and once again successfully finds a lower slope region to climb down.

\section{CONCLUSION}

In this work we have presented a local mapping approach for off-road ground vehicles. Our system takes in vehicle odometry and lidar data to produce a local 3D voxel map around the vehicle and outputs both the completed voxel map and a series of 2D maps for local planning. It successfully extracts both hard and soft positive obstacles, negative obstacles, extracts a height map of the terrain, and provides roughness and slope metrics. This is accomplished at a sufficient rate for online operation, 10~Hz on our laptop. We demonstrated the system's effectiveness in multiple scenarios at up to 4.5~m/s in autonomous operation and 12~m/s in manual operation.

\balance



\clearpage

\bibliographystyle{IEEEtran}
\bibliography{IEEEabrv,IEEEexample}

\begin{thebibliography}{10}
\providecommand{\url}[1]{#1}
\csname url@rmstyle\endcsname
\providecommand{\newblock}{\relax}
\providecommand{\bibinfo}[2]{#2}
\providecommand\BIBentrySTDinterwordspacing{\spaceskip=0pt\relax}
\providecommand\BIBentryALTinterwordstretchfactor{4}
\providecommand\BIBentryALTinterwordspacing{\spaceskip=\fontdimen2\font plus
\BIBentryALTinterwordstretchfactor\fontdimen3\font minus
  \fontdimen4\font\relax}
\providecommand\BIBforeignlanguage[2]{{%
\expandafter\ifx\csname l@#1\endcsname\relax
\typeout{** WARNING: IEEEtran.bst: No hyphenation pattern has been}%
\typeout{** loaded for the language `#1'. Using the pattern for}%
\typeout{** the default language instead.}%
\else
\language=\csname l@#1\endcsname
\fi
#2}}

\bibitem{probobalistic_robotics}
S.~Thrun, W.~Burgard, and D.~Fox, \emph{Probabilistic Robotics (Intelligent
  Robotics and Autonomous Agents)}.\hskip 1em plus 0.5em minus 0.4em\relax The
  MIT Press, 2005.

\bibitem{occupancy_grid}
A.~Elfes, ``Using occupancy grids for mobile robot perception and navigation,''
  \emph{Computer}, vol.~22, no.~6, pp. 46--57, 1989.

\bibitem{grid_map_1}
G.~Grisetti, C.~Stachniss, and W.~Burgard, ``Improved techniques for grid
  mapping with rao-blackwellized particle filters,'' \emph{IEEE Transactions on
  Robotics}, vol.~23, no.~1, pp. 34--46, 2007.

\bibitem{multilayer_obs}
D.~V. Lu, D.~Hershberger, and W.~D. Smart, ``Layered costmaps for
  context-sensitive navigation,'' in \emph{2014 IEEE/RSJ International
  Conference on Intelligent Robots and Systems}, 2014, pp. 709--715.

\bibitem{multilayer_sensors}
S.~Mentasti and M.~Matteucci, ``Multi-layer occupancy grid mapping for
  autonomous vehicles navigation,'' in \emph{2019 AEIT International Conference
  of Electrical and Electronic Technologies for Automotive (AEIT AUTOMOTIVE)},
  2019, pp. 1--6.

\bibitem{my_planning_paper}
T.~Overbye and S.~Saripalli, ``Fast local planning and mapping in unknown
  off-road terrain,'' in \emph{2020 IEEE International Conference on Robotics
  and Automation (ICRA)}, 2020, pp. 5912--5918.

\bibitem{drivable_learning_1}
B.~Gao, A.~Xu, Y.~Pan, X.~Zhao, W.~Yao, and H.~Zhao, ``Off-road drivable area
  extraction using 3d lidar data,'' 06 2019, pp. 1505--1511.

\bibitem{drivable_learning_2}
B.~Suger, B.~Steder, and W.~Burgard, ``Traversability analysis for mobile
  robots in outdoor environments: A semi-supervised learning approach based on
  3d-lidar data,'' in \emph{2015 IEEE International Conference on Robotics and
  Automation (ICRA)}, 2015, pp. 3941--3946.

\bibitem{multilevel_surface}
R.~Triebel, P.~Pfaff, and W.~Burgard, ``Multi-level surface maps for outdoor
  terrain mapping and loop closing,'' in \emph{2006 IEEE/RSJ International
  Conference on Intelligent Robots and Systems}, 2006, pp. 2276--2282.

\bibitem{voxel_1}
Y.~Roth-Tabak and R.~Jain, ``Building an environment model using depth
  information,'' \emph{Computer}, vol.~22, no.~6, pp. 85--90, 1989.

\bibitem{voxel_occupancy}
P.~Payeur, P.~Hebert, D.~Laurendeau, and C.~Gosselin, ``Probabilistic octree
  modeling of a 3d dynamic environment,'' in \emph{Proceedings of International
  Conference on Robotics and Automation}, vol.~2, 1997, pp. 1289--1296 vol.2.

\bibitem{octree}
\BIBentryALTinterwordspacing
D.~Meagher, ``Geometric modeling using octree encoding,'' \emph{Computer
  Graphics and Image Processing}, vol.~19, no.~2, pp. 129--147, 1982. [Online].
  Available:
  \url{https://www.sciencedirect.com/science/article/pii/0146664X82901046}
\BIBentrySTDinterwordspacing

\bibitem{octomap}
\BIBentryALTinterwordspacing
A.~Hornung, K.~M. Wurm, M.~Bennewitz, C.~Stachniss, and W.~Burgard,
  ``{OctoMap}: An efficient probabilistic {3D} mapping framework based on
  octrees,'' \emph{Autonomous Robots}, 2013, software available at
  \url{http://octomap.github.com}. [Online]. Available:
  \url{http://octomap.github.com}
\BIBentrySTDinterwordspacing

\bibitem{jittree}
M.~Labsch{\"u}tz, S.~Bruckner, M.~E. Gr{\"o}ller, M.~Hadwiger, and P.~Rautek,
  ``Jittree: A just-in-time compiled sparse gpu volume data structure,''
  \emph{IEEE Transactions on Visualization and Computer Graphics (Proceedings
  IEEE Scientific Visualization 2015)}, vol.~22, no.~1, pp. 1025--1034, 2016.

\bibitem{octree_gpu}
S.~Laine and T.~Karras, ``Efficient sparse voxel octrees,'' \emph{IEEE
  Transactions on Visualization and Computer Graphics}, vol.~17, no.~8, pp.
  1048--1059, 2011.

\bibitem{skimap}
D.~De~Gregorio and L.~Di~Stefano, ``Skimap: An efficient mapping framework for
  robot navigation,'' 04 2017.

\bibitem{virtual_surfaces}
T.~Hines, K.~Stepanas, F.~Talbot, I.~Sa, J.~Lewis, E.~Hernandez, N.~Kottege,
  and N.~Hudson, ``Virtual surfaces and attitude aware planning and behaviours
  for negative obstacle navigation,'' \emph{IEEE Robotics and Automation
  Letters}, vol.~PP, pp. 1--1, 03 2021.

\bibitem{loam}
J.~Zhang and S.~Singh, ``Loam : Lidar odometry and mapping in real-time,''
  \emph{Robotics: Science and Systems Conference (RSS)}, pp. 109--111, 01 2014.

\bibitem{my_optimization_paper}
T.~Overbye and S.~Saripalli, ``Path optimization for ground vehicles in
  off-road terrain,'' in \emph{2021 IEEE International Conference on Robotics
  and Automation (ICRA)}, 2021.

\bibitem{ilqr}
A.~Nagariya and S.~Saripalli, ``An iterative lqr controller for off-road and
  on-road vehicles using a neural network dynamics model,'' in \emph{2020 IEEE
  Intelligent Vehicles Symposium (IV)}, 2020, pp. 1740--1745.

\end{thebibliography}

\end{document}